\documentclass[11pt]{article}

\usepackage[preprint]{acl}

\usepackage{times}
\usepackage{latexsym}

\usepackage[T1]{fontenc}
\usepackage[utf8]{inputenc}

\usepackage{microtype}
\usepackage{tabularx}
\usepackage{inconsolata}

\usepackage{graphicx}

\usepackage{amsmath, amssymb}
\usepackage{booktabs}
\usepackage{multirow}

\usepackage{algorithm}
\usepackage{algpseudocode}
\usepackage{pifont} % Add this in your preamble for check/cross marks

\newcommand{\cmark}{\ding{51}} % check mark
\newcommand{\xmark}{\ding{55}} % cross mark
\usepackage{bbm}
\usepackage{tikz}
\newlength{\barwidth}
\usepackage{lipsum}

\usepackage{hyperref}
\usepackage{CJKutf8}

\title{UXBench: Benchmarking User Experience in AI Assistants}

\author{
\textbf{Mengze Hong$^{1,2}$, Xia Zeng$^{2}$, Zeyang Lei$^{2*}$, Sheng Wang$^{2}$, Chen Jason Zhang$^{1*}$, Di Jiang$^{1}$} \\
\textbf{Taiming Fu$^{2}$, Jinfeng Huang$^{2}$, Mengqiao Liu$^{2}$, Qinghe Chang$^{2}$, Haosheng Zou$^{2}$} \\
\textbf{Qiongyi Zhou$^{2}$, Sijun He$^{2}$, Simonjmdeng$^{2}$, Minlong Peng$^{2}$, Di Liang$^{2}$\thanks{Corresponding authors.}} \vspace{0.3em} \\
$^{1}$Hong Kong Polytechnic University, $^{2}$Yuanbao Team, Tencent
}

\begin{document}
\maketitle

\begin{abstract}
As AI assistants serve millions of users daily, evaluating user experience (UX) beyond general model capability has become increasingly important. We present \textbf{UXBench}, the first user-centric benchmark grounded in real user feedback signals for evaluating preference alignment and dialogue generation. The benchmark consists of three interconnected tasks, UX Judge, UX Eval, and UX Recovery, with 7,400 test instances extracted from over 70K interaction logs of a mainstream Chinese AI assistant. The dataset closely reflects real user distributions, covering 8 scenarios, 83 domains, and diverse failure patterns that pose severe challenges. Extensive experiments on 26 frontier language models provide novel insights into how well models perceive user experience and how improvements in model capability contribute to better dialogue engagement. Through analyses of model behavior and performance gaps, we document six important findings, demonstrating that user feedback prediction is a learnable capability and revealing different aspects that influence user experience. UXBench establishes a new evaluation landscape and calls for greater attention to tailored UX optimization, contributing toward a user-centric scaling law for the development of successful AI assistants. The full project is released at https://github.com/mengze-hong/UXBench.
\end{abstract}

\begin{flushright}
\textit{What makes a good product? \\Built on functionality, won by experience.}
\end{flushright}

\section{Introduction}

The long-term success of AI products relies on retaining active users by delivering a positive user experience (UX). Existing LLM benchmarks primarily evaluate foundation model capabilities such as reasoning, knowledge, and instruction following \citep{mohammadi2025evaluation, guo-etal-2025-sdbench}, while largely neglecting user-perceived utility and interaction quality. It remains unclear whether stronger models under existing evaluations necessarily deliver better real-world user experiences, as improvements on traditional benchmarks may translate into only marginal or imperceptible gains for end users who mostly engage in casual information search and chit-chat interactions \citep{chatterji2025people}. This motivates the need for a rigorous UX-centric benchmark grounded in realistic user feedback and interaction behavior to better understand and reassess progress in AI assistants.

Developing a dedicated benchmark for UX presents three fundamental challenges. First, existing UX research is largely rooted in human-computer interaction (HCI), where experience metrics are often assessed through studies involving human participants that are difficult to scale \citep{10.1145/3706598.3714045,  10.1145/3772318.3791724}. Second, many evaluation schemes quantify UX through fine-grained dimensions \citep{mahmud2025evaluating, li2026prefixunderstandadaptuser}, which, while interpretable, cannot be exhaustively defined or fully operationalized. Intuitively, users do not explicitly deliberate before clicking the ``unlike'' button; rather, it is an immediate reaction to unsatisfactory responses \citep{ntoa2025usability, 10.1145/3706598.3714045}. Meanwhile, computational approaches such as arena-style benchmarks \citep{chiang2024chatbot} focus primarily on pairwise model comparisons, which may not align with behavioral signals from real-world interactions, where users typically engage with only one response at a time.

\begin{table*}[!t]
\centering \footnotesize
\resizebox{\textwidth}{!}{
\begin{tabular}{lccccccc}
\toprule
Benchmark & Source & Scale (N) & Avg Turns & Signal & Judge & UX-driven & User Complains \\
\midrule
MT-Bench~\citep{NEURIPS2023_91f18a12} & Curated & 80 & 2 & Rubric & LLM & \xmark & \xmark \\
AlpacaEval~\citep{alpaca_eval} & Synth & 805 & 1 & Pairwise & LLM & \xmark & \xmark \\
Chatbot Arena~\citep{chiang2024chatbot} & Logs & Dynamic & 1 & Voting & Human & \xmark & \xmark \\
Arena-Hard~\citep{li2025crowdsourced} & Curated & 500 & 1 & Pairwise & LLM & \xmark & \xmark \\
WildBench~\citep{lin2025wildbench} & Logs & 1,024 & 2-3 & Rubric & LLM & \xmark & \xmark \\
\midrule
\textbf{UXBench (Ours)} & \textbf{Logs} & \textbf{7,400} & \textbf{5.29} & \textbf{Feedback} & \textbf{GRM} & \textbf{\cmark} & \textbf{\cmark} \\
\bottomrule
\end{tabular}}
\caption{Comparison of UXBench with existing dialogue-system evaluation benchmarks.}
\label{tab:benchmark_comparison}
\end{table*}

In this paper, we introduce \textbf{UXBench}, the first user-centric benchmark designed for automated user experience assessment in AI assistants. Built on large-scale human--AI interaction logs, we propose three interconnected tasks that enable faithful and scalable evaluation of user-perceived utility, revealing key empirical insights for advancing UX benchmarking and model optimization. The main contributions include:

\begin{enumerate}

\item \textbf{User Experience Modeling}: We leverage user feedback signals from real-time interactions as ground-truth UX labels, turning subjective user experience into an observable and scalable modeling target, and identify three key signal categories that support UX-oriented evaluation and optimization at scale.

\item \textbf{Evaluation Framework}: We introduce UXBench, a user-centric benchmark comprising 7,400 test cases sampled from over 70K real interaction logs for evaluating LLMs in user feedback prediction and response generation. A rigorous multi-stage data pipeline enables automated discovery of reliable user feedback signals.

\item \textbf{Actionable Insights}: Extensive evaluation on 26 frontier LLMs across three UX tasks reveals six key findings on model performance, failure modes, systematic biases, and scaling trends. A trained reward model further achieves state-of-the-art performance in predicting user feedback, motivating model-based UX optimization.

\end{enumerate}

\begin{figure*}[!t]
    \centering
    \includegraphics[width=1\linewidth]{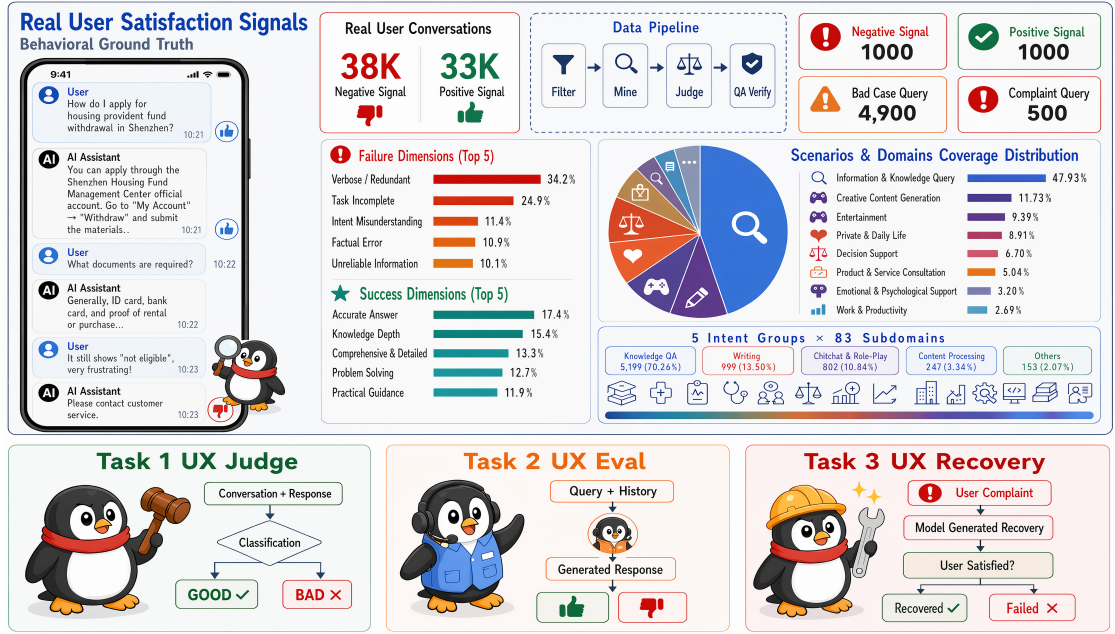}
    \vspace{-2em}
    \caption{Overview of UXBench, including real-user data sources, data processing pipeline, key statistics across success and failure dimensions, scenario and domain coverage, and an illustration of the three proposed tasks.}
    \label{fig:overview_of_uxbench}
\end{figure*}

\section{Related Work}

\subsection{Benchmarking Dialogue Systems}

The rise of personalized AI assistants has shifted evaluation beyond traditional QA-based benchmarks toward open-ended dialogue interactions \citep{10.1145/3771090}, in which evaluation protocols primarily follow pairwise, rubric-based, and pointwise paradigms. As shown in Table \ref{tab:benchmark_comparison}, pairwise and rubric-guided methods are widely used for preference modeling and structured evaluation \citep{liu-etal-2023-g, zhang2026rubricbench}, with representative benchmarks such as URS \citep{wang-etal-2024-user,wang2025not} and WildBench \citep{lin2025wildbench}. More recently, pointwise evaluation has gained attention for improved robustness and interpretability, as shown in PrefIx that employs multi-LLM judges across seven Likert-scale dimensions \citep{li2026prefixunderstandadaptuser,liang2019asynchronous}, and is generally more resistant to positional bias and external manipulation \citep{tripathi2025pairwise}. However, existing benchmarks remain limited by poorly aligned LLM judges and insufficient coverage of failure-prone user queries, resulting in an incomplete understanding of user-perceived response quality.

\subsection{Benchmarking User Experience}

User experience is a central research agenda in HCI, reflecting how effectively, efficiently, and satisfactorily users interact with systems \citep{ntoa2025usability}. Traditional methods such as the User Experience Questionnaire (UEQ) provide structured self-reported feedback but struggle to capture the dynamic, interactive nature of AI systems \citep{10.1145/3706598.3714045,lin2026parameter}. Specialized frameworks, such as CASUX \citep{faruk2025introducing}, have been proposed for dialogue agents to combine qualitative analysis with quantitative UX metrics. However, these approaches remain largely static and depend heavily on human annotators and manually defined evaluation rubrics, limiting their ability to capture the complexity and diversity of real user behavior \citep{mahmud2025evaluating}. UXBench departs from this tradition by leveraging real user feedback signals as direct ground-truth annotations, and by training user-aligned reward models for fully automated evaluation. This reduces reliance on subjective post-hoc judgments and enables more faithful, scalable, and user-aligned model assessment.

\subsection{Reward Modeling}

Reward models (RMs) are central to modern LLM alignment with user feedback \citep{NEURIPS2022_b1efde53}, with an increasing number of studies emphasizing their training and evaluation \citep{peng-etal-2026-wildreward, ICLR2026_ea4fe0a5}. Early scalar RMs have gradually evolved into generative reward models (GRMs) that produce natural language critiques alongside scalar judgments \citep{pmlr-v267-wang25ad}. On the other hand, the quality of RMs has been extensively evaluated through public benchmarks such as RewardBench \citep{lambert-etal-2025-rewardbench,xue2026supervised,liu2025structural} and PPE \citep{ICLR2025_2e01083b}, both of which primarily rely on pairwise ranking against human annotations. In this paper, we leverage RM as a standalone instrument for scalable dialogue evaluation. Unlike conventional LLM-as-a-judge methods \citep{li-etal-2025-generation}, which rely on prompting or rubric-based evaluation and remain susceptible to biases from imperfect preference alignment \citep{chen-etal-2024-humans,liang2019adaptive, ICLR2025_fdca08d3}, GRMs trained on authentic user preference signals may provide more human-aligned evaluation.

\section{UXBench}

\subsection{Overview}

UXBench is designed to answer three core research questions: (1) how well automated LLM judges predict real user feedback to an AI-generated response; (2) whether frontier LLMs can generate high-quality responses for failure-prone user queries; and (3) how progress in model capability improvements translates into measurable UX gains. Rather than decomposing user experience into isolated dimensions, we formulate positive and negative user feedback as a unified behavioral signal and construct challenging, quality-audited test sets from real interaction logs (Section~\ref{sec:data signal}). Based on this formulation, UXBench defines three interconnected tasks (Section~\ref{sec:task formulation}), which progressively assess model capabilities to align with user perception, generate satisfying responses, and recover from service failures. Figure~\ref{fig:overview_of_uxbench} summarizes the data sources, key features, and task design.

To excel in this benchmark, models require not only fundamental reasoning and retrieval capabilities but, more importantly, a user-aligned understanding of what constitutes a good interaction,  appropriate dialogue strategies, and sufficient emotional intelligence. Ideally, strong performance on UXBench would directly translate into a better real-world user experience for end users.

\subsection{UX Modeling with Feedback Signals}
\label{sec:data signal}

\begin{figure}
    \centering
    \includegraphics[width=1\linewidth]{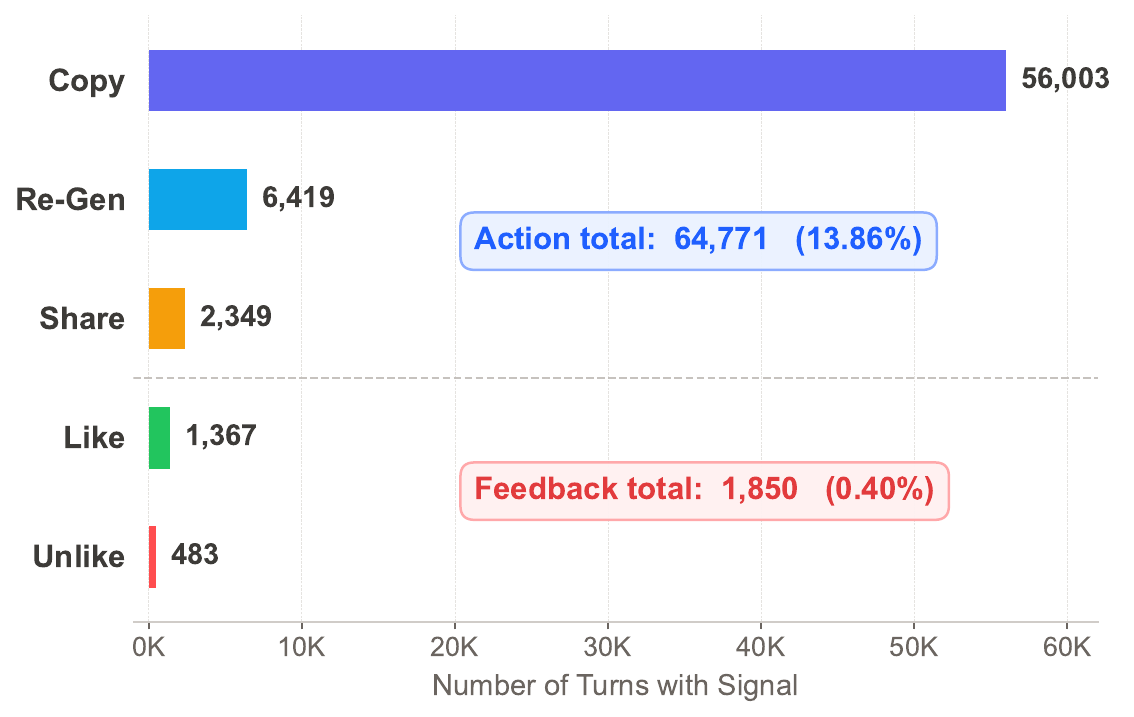}
\caption{Distribution of explicit feedback signals from 400K randomly sampled dialogue turns.}
    \label{fig:signal_distribution}
\end{figure}

A key bottleneck in user experience evaluation is the lack of reliable ground-truth labels. Existing approaches often rely on human interpretation or post-hoc annotation, making the process subjective and susceptible to annotator bias \citep{weber-genzel-etal-2024-varierr}. To address this challenge, we leverage feedback signals from real-time user interactions and treat them as the most directly observable indicators of response quality \citep{liu-etal-2025-user}. Modern AI systems commonly provide explicit feedback mechanisms, such as \texttt{like} and \texttt{dislike} buttons. However, providing such feedback contributes negligibly to task completion and offers little immediate benefit to users, making explicit labels reliable but inherently sparse for scalable UX modeling. In a random sample of 400K interaction turns (Figure~\ref{fig:signal_distribution}), we find that such feedback signals are extremely sparse compared with action-based behavioral signals.

Action-based signals remain inherently ambiguous and prone to misinterpretation. For example, sharing a response may reflect either satisfaction or an intention to complain. Thus, we complement explicit feedback with \textbf{implicit negative feedback} derived from inter-turn user behaviors. Prior work has explored shallow implicit signals, including user input content \citep{Zhang_Li_Cao_Liu_2018} and statistics such as input length and response latency \citep{pang-etal-2024-leveraging}. By analyzing turns surrounding explicit dislikes, we derive two rule-based heuristics for implicit negative feedback. First, \texttt{Skim} captures rapid follow-up without fully engaging with the response. Let $\Delta\tau_{t+1}=\tau_{t+1}^{u}-\tau_t^{a}$ denote the time to the next user input and $|r_t^{a}|$ the response length. A turn is flagged when
\begin{equation}
\frac{\Delta\tau_{t+1}}{|r_t^{a}|}<0.15
\quad \text{and} \quad
\Delta\tau_{t+1}<30\text{s},
\end{equation}
with thresholds calibrated on skim cases preceding explicit dislikes. Second, \texttt{follow-up dissatisfaction} captures subsequent \texttt{Complain}/\texttt{Dismiss} turns via fixed dissatisfaction regexes and \texttt{Repeat}/\texttt{Rephrase} turns via high lexical similarity to the preceding query. A locally deployed LLM further verifies heuristic-matched instances for quality assurance. Across 40K sampled dialogue sessions, we identify six negative signals: Repeat (10,337), Rephrase (4,773), Skim (756), Complain (199), Dismiss (187), and explicit Dislike (65; 0.16\%), revealing diverse manifestations of user dissatisfaction.

\subsection{Task Formulation}
\label{sec:task formulation}

\paragraph{Task 1: UX Judge.} We first evaluate the model's ability to predict user feedback from dialogue interactions. Formally, let
\begin{equation}
\mathcal{D} = \left\{(h_i, q_i, r_i, y_i)\right\}_{i=1}^{N}, \quad
y_i \in \left\{y_i^{+}, y_i^{-}\right\}
\end{equation}
denote a collection of dialogue trajectories, where $h_i$ is the dialogue history, $q_i$ is the user query, $r_i$ is the assistant response, and $y_i^{+}/y_i^{-}$ are the corresponding user feedback signals. A language model $\mathcal{M}_{\theta}$ performs binary UX judgment: 
$\mathcal{M}_{\theta}(h_i, q_i, r_i) \mapsto \hat{y}_i \in \{0,1\}$, predicting latent user satisfaction from interaction content, with accuracy and class-wise recall reported. Although binary prediction is less expressive than rubric-based ordinal ratings, it is motivated by stronger alignment with human judgments under 0--1 loss compared with ordinal or Likert-scale regression settings \citep{hong-etal-2025-dial}, while directly reflecting real user feedback behavior and providing natural supervision for training and evaluation. In Section~\ref{sec:comparison of formulations}, we compare binary judgment with several rubric-based formulations, supporting this design choice as a practical first step toward quantifying user experience.

\paragraph{Task 2: UX Eval.} 
The second task evaluates the model-generated responses to determine whether they lead to a positive user experience. Let $\mathcal{D} = \left\{(h_i, q_i, y_i^{-})\right\}_{i=1}^{N}$ denote dialogue sessions extracted exclusively from interactions where the original AI response received explicit negative feedback, forming a dataset intentionally skewed toward failure-prone, hard user cases rather than average interactions with limited discriminative value. Given a generated response $\hat{r}_i$, a reward model $\mathcal{R}_{\phi}$ predicts a UX score $s_i=\mathcal{R}_{\phi}(h_i,q_i,\hat{r}_i)\in[0,1]$, which is converted into a binary preference label $\hat{y}_i=\mathbbm{1}[s_i\geq0.5]$ to report the overall percentage of positive responses:
\begin{equation}
\mathrm{Good\%} = \frac{1}{N}\sum_{i=1}^{N}\hat{y}_i.
\end{equation}

\paragraph{Task 3: UX Recovery.} The third task extends UX Eval to long-tail, failure-critical scenarios by evaluating whether a model can recover user satisfaction after an explicit interaction failure. Each instance is defined as $d_i = (h_i, c_i)$, where $h_i$ denotes the dialogue history and $c_i$ is a user complaint. From a random sample of 2K complaints, we observe that over 83\% of interactions terminate immediately after a poorly drafted recovery response, highlighting the importance of effective recovery strategies for user retention. We follow the same evaluation protocol as Task 2, while UX Recovery specifically measures whether a model can repair failed interactions through appropriate dialogue strategies, emotional awareness, and corrective behavior under realistic user complaints.

\subsection{Dataset Construction Pipeline}
\label{sec:data_pipeline}

UXBench is constructed from 70K+ real user-AI interactions extracted from a mainstream Chinese AI assistant that supports chit-chat, role-play, and information-grounded question answering. Ethical data collection and user privacy are treated as top priorities. All data is collected with explicit user consent through an opt-in experience program, in which a group of users voluntarily authorize the use of their interaction logs for research and service improvement. The pipeline is formulated as a fully automated end-to-end transformation from raw dialogue logs to human-interpretable test cases labeled with feedback signals. A dialogue session \(\mathbf{D}_i = \{x_1^{u}, x_1^{a}, \ldots, x_T^{u}, x_T^{a}\}\), first undergoes de-identification and desensitization \(\tilde{\mathbf{D}}_i = \phi_{\text{anon}}(\mathbf{D}_i)\) using rule-based and locally deployed LLM \citep{yang-etal-2025-robust}. A deterministic gate \(\phi_{\text{gate}}\) then performs data cleaning and removes unsafe cases involving sexual, violent, or other harmful content, after which behavioral feedback signals are derived as \(\mathbf{s}_i = \phi_{\text{signal}}(\tilde{\mathbf{D}}_i)\) through pre-defined heuristics.

Given pre-processed dialogue sessions, a multi-agent system constructs test cases using locally deployed Hunyuan and DeepSeek models. A Miner model first proposes failure candidates \(b \in \mathcal{C}_i\) covering diverse failure modes, while an independent Judge model assigns quality scores \(q_b \in [1,5]^5\) across five dimensions: completeness, credibility, representativeness, severity, and signal fidelity. Candidates are retained if \(\bar{q}_b \ge 4.0\) and \(\min(q_b) \ge 3\). Finally, the resulting dataset undergoes LLM-based quality assurance and human expert verification prior to application.

\begin{table}[t]
\centering
\small
\setlength{\tabcolsep}{3.5pt}
\resizebox{\columnwidth}{!}{
\begin{tabular}{@{}lcccc@{}}
\toprule
& \textbf{Task 1} & \textbf{Task 1} & \textbf{Task 2} & \textbf{Task 3} \\
& \textbf{BAD} & \textbf{GOOD} & \textbf{Eval} & \textbf{Recovery} \\
\midrule
Instances & 1,000 & 1,000 & 4,900 & 500 \\
Multi-turn (\%) & 71.3 & 71.4 & 71.7 & 98.0 \\
Average turns & 5.1  & 5.2 & 5.2 & 6.2 \\
Mean query length (chars) & 26.1 & 38.9 & 31.7 & 85.9 \\
Mean response length (chars) & 836 & 748 & -- & --\\
\bottomrule
\end{tabular}}
\caption{Key statistics of the UXBench dataset.}
\label{tab:dataset_stats}
\end{table}

\paragraph{Data Statistics.}

Table~\ref{tab:dataset_stats} reports key statistics of UXBench across three tasks, comprising 7,400 test cases spanning 8 interaction scenarios and 83 domains. Since the data are randomly sampled from online interaction logs, the scenario distribution closely resembles real-world user interactions, ensuring that the benchmark serves as a reliable proxy across diverse user queries. The dataset mainly consists of multi-turn interactions, enabling evaluation under long-context settings. A key feature of UXBench is its structured success and failure taxonomy: negative instances cover 10 failure types, with verbosity being the most common issue, while positive instances span 8 success types, dominated by accurate answering (see Appendix~\ref{appendix:taxonomy of dimensions}).

% Main Result Table (Task 1 + Task 2 + Task 3)
% Requires: \usepackage{tikz,booktabs}
\ifdefined\barwidth\else\newlength{\barwidth}\fi
\setlength{\barwidth}{35pt}
\providecommand{\progressbargood}[1]{%
  \begin{tikzpicture}[baseline=-0.1ex]%
    \fill[gray!20] (0,0) rectangle (\barwidth,5pt);%
    \fill[blue!45] (0,0) rectangle (#1/100*\barwidth,5pt);%
  \end{tikzpicture}%
}
\providecommand{\progressbarbad}[1]{%
  \begin{tikzpicture}[baseline=-0.1ex]%
    \fill[gray!20] (0,0) rectangle (\barwidth,5pt);%
    \fill[orange!55] (0,0) rectangle (#1/100*\barwidth,5pt);%
  \end{tikzpicture}%
}
\providecommand{\goldmedal}{%
  \tikz[baseline=-0.8ex]{\node[circle, fill=yellow!70!orange, draw=orange!80!black, line width=0.4pt, text=white, font=\tiny\bfseries, inner sep=1pt]{1};}%
}
\providecommand{\silvermedal}{%
  \tikz[baseline=-0.8ex]{\node[circle, fill=gray!40, draw=gray!70, line width=0.4pt, text=white, font=\tiny\bfseries, inner sep=1pt]{2};}%
}
\providecommand{\bronzemedal}{%
  \tikz[baseline=-0.8ex]{\node[circle, fill=orange!70, draw=orange!90!red, line width=0.4pt, text=white, font=\tiny\bfseries, inner sep=1pt]{3};}%
}

\begin{table*}[!t]
\centering\footnotesize
\setlength{\tabcolsep}{4pt}
\resizebox{\textwidth}{!}{
\begin{tabular}{l r r @{\hspace{2pt}} l r @{\hspace{2pt}} l r @{\hspace{2pt}} l r @{\hspace{2pt}} l}
\toprule
\multirow{2}{*}{\textbf{Model}}
  & \multicolumn{5}{c}{\textbf{Task 1: UX Judge}}
  & \multicolumn{2}{c}{\textbf{Task 2: UX Eval}}
  & \multicolumn{2}{c}{\textbf{Task 3: UX Recovery}} \\
\cmidrule(lr){2-6}\cmidrule(lr){7-8}\cmidrule(lr){9-10}
  & Acc\% & \multicolumn{2}{c}{Good-Acc\%} & \multicolumn{2}{c}{Bad-Acc\%}
  & \multicolumn{2}{c}{Good\%} & \multicolumn{2}{c}{Good\%} \\
\midrule

\multicolumn{10}{l}{\textit{Google Gemini}} \\
\quad Gemini 2.5 Flash       & 59.4\% & 97.5\% & \progressbargood{97.5} & 21.2\% & \progressbarbad{21.2} & 36.1\% & \progressbargood{36.1} & 4.6\% & \progressbargood{4.6} \\
\quad Gemini 2.5 Pro         & 62.8\% & 96.8\% & \progressbargood{96.8} & 28.7\% & \progressbarbad{28.7} & 50.8\% & \progressbargood{50.8} & 4.0\% & \progressbargood{4.0} \\
\quad Gemini 3.0 Flash       & 67.4\% & 97.7\% & \progressbargood{97.7} & 37.1\% & \progressbarbad{37.1} & 52.7\% & \progressbargood{52.7} & \silvermedal\,12.7\% & \progressbargood{12.7} \\
\quad Gemini 3.1 Pro         & 70.4\% & 91.6\% & \progressbargood{91.6} & 49.3\% & \progressbarbad{49.3} & \goldmedal\,\textbf{57.1\%} & \progressbargood{57.1} & 9.2\% & \progressbargood{9.2} \\
\midrule
\multicolumn{10}{l}{\textit{OpenAI}} \\
\quad GPT-5 mini             & 65.2\% & 93.4\% & \progressbargood{93.4} & 36.9\% & \progressbarbad{36.9} & 24.0\% & \progressbargood{24.0} & 3.6\% & \progressbargood{3.6} \\
\quad GPT-5                  & 72.9\% & 89.5\% & \progressbargood{89.5} & 56.2\% & \progressbarbad{56.2} & 34.7\% & \progressbargood{34.7} & 6.9\% & \progressbargood{6.9} \\
\quad GPT-5.1                & 72.5\% & 94.8\% & \progressbargood{94.8} & 50.1\% & \progressbarbad{50.1} & 37.1\% & \progressbargood{37.1} & 7.4\% & \progressbargood{7.4} \\
\quad GPT-5.2                & \silvermedal\,75.0\% & 85.0\% & \progressbargood{85.0} & \textbf{65.1\%} & \progressbarbad{65.1} & 30.8\% & \progressbargood{30.8} & 5.4\% & \progressbargood{5.4} \\
\quad GPT-5.5                & \bronzemedal\,74.2\% & 92.7\% & \progressbargood{92.7} & 55.7\% & \progressbarbad{55.7} & 41.2\% & \progressbargood{41.2} & 9.9\% & \progressbargood{9.9} \\
\midrule
\multicolumn{10}{l}{\textit{Anthropic}} \\
\quad Claude Sonnet 4.5      & 69.3\% & 89.6\% & \progressbargood{89.6} & 49.0\% & \progressbarbad{49.0} & 36.0\% & \progressbargood{36.0} & 9.0\% & \progressbargood{9.0} \\
\quad Claude Opus 4.5        & 66.7\% & 96.7\% & \progressbargood{96.7} & 36.7\% & \progressbarbad{36.7} & 37.9\% & \progressbargood{37.9} & 9.4\% & \progressbargood{9.4} \\
\quad Claude Opus 4.6        & 72.0\% & 92.6\% & \progressbargood{92.6} & 51.5\% & \progressbarbad{51.5} & 44.3\% & \progressbargood{44.3} & \goldmedal\,\textbf{12.8\%} & \progressbargood{12.8} \\
\quad Claude Opus 4.7        & \goldmedal\,\textbf{75.3\%} & 89.1\% & \progressbargood{89.1} & 61.5\% & \progressbarbad{61.5} & 44.5\% & \progressbargood{44.5} & \bronzemedal\,12.4\% & \progressbargood{12.4} \\
\midrule
\multicolumn{10}{l}{\textit{DeepSeek}} \\
\quad DeepSeek R1            & 58.5\% & 98.7\% & \progressbargood{98.7} & 18.3\% & \progressbarbad{18.3} & 39.5\% & \progressbargood{39.5} & 5.8\% & \progressbargood{5.8} \\
\quad DeepSeek V3            & 55.6\% & \textbf{99.7\%} & \progressbargood{99.7} & 11.6\% & \progressbarbad{11.6} & 35.9\% & \progressbargood{35.9} & 3.6\% & \progressbargood{3.6} \\
\quad DeepSeek V3.2          & 64.5\% & 95.7\% & \progressbargood{95.7} & 33.3\% & \progressbarbad{33.3} & 41.2\% & \progressbargood{41.2} & 7.0\% & \progressbargood{7.0} \\
\quad DeepSeek V4 Pro        & 64.5\% & 97.4\% & \progressbargood{97.4} & 31.7\% & \progressbarbad{31.7} & 49.7\% & \progressbargood{49.7} & 11.0\% & \progressbargood{11.0} \\
\midrule
\multicolumn{10}{l}{\textit{ByteDance}} \\
\quad Doubao Seed 1.6        & 57.6\% & 99.1\% & \progressbargood{99.1} & 16.2\% & \progressbarbad{16.2} & 36.8\% & \progressbargood{36.8} & 6.8\% & \progressbargood{6.8} \\
\quad Doubao Seed 2.0 Lite   & 57.4\% & 98.7\% & \progressbargood{98.7} & 16.0\% & \progressbarbad{16.0} & 46.3\% & \progressbargood{46.3} & 10.4\% & \progressbargood{10.4} \\
\quad Doubao Seed 2.0 Pro    & 60.8\% & 98.8\% & \progressbargood{98.8} & 22.9\% & \progressbarbad{22.9} & 48.7\% & \progressbargood{48.7} & 10.8\% & \progressbargood{10.8} \\
\midrule
\multicolumn{10}{l}{\textit{Others}} \\
\quad Hunyuan 3 Preview    & 64.3\% & 95.6\% & \progressbargood{95.6} & 33.1\% & \progressbarbad{33.1} & 48.8\% & \progressbargood{48.8} & 7.6\% & \progressbargood{7.6} \\
\quad GLM-5                  & 66.7\% & 96.9\% & \progressbargood{96.9} & 36.4\% & \progressbarbad{36.4} & \bronzemedal\,53.0\% & \progressbargood{53.0} & 10.4\% & \progressbargood{10.4} \\
\quad GLM-5.1                & 68.5\% & 96.1\% & \progressbargood{96.1} & 40.9\% & \progressbarbad{40.9} & \silvermedal\,56.6\% & \progressbargood{56.6} & 11.2\% & \progressbargood{11.2} \\
\quad Kimi K2.5              & 64.8\% & 96.8\% & \progressbargood{96.8} & 32.7\% & \progressbarbad{32.7} & 50.3\% & \progressbargood{50.3} & 11.2\% & \progressbargood{11.2} \\
\quad Kimi K2.6              & 68.7\% & 96.1\% & \progressbargood{96.1} & 41.2\% & \progressbarbad{41.2} & 52.3\% & \progressbargood{52.3} & 11.4\% & \progressbargood{11.4} \\
% \quad MiniMax M2.5           & 61.3\% & 97.8\% & \progressbargood{97.8} & 24.8\% & \progressbarbad{24.8} & 28.0\% & \progressbargood{28.0} & 7.6\% & \progressbargood{7.6} \\
\quad Qwen3.6-Plus           & 65.8\% & 96.8\% & \progressbargood{96.8} & 34.9\% & \progressbarbad{34.9} & 52.3\% & \progressbargood{52.3} & 12.0\% & \progressbargood{12.0} \\
\midrule
\multicolumn{10}{l}{\textit{Trained Model}} \\
\quad \textbf{Pointwise GRM (ours)} & \textbf{77.2\%} & \textbf{82.1\%} & \progressbargood{82.1} & \textbf{72.4\%} & \progressbarbad{72.4} & -- & & -- & \\
\bottomrule
\end{tabular}}
\caption{%
  UXBench results across 26 frontier LLMs and trained Pointwise GRM, grouped by model family.
}
\label{tab:main_result}
\end{table*}

\section{Results and Discussion}

In this section, we progressively analyze LLM capabilities from user feedback prediction to response generation, complemented by key ablation studies for each task. Complete experimental results are reported in Appendix~\ref{appendix:full-results}. Since a key research question is to understand how UX evolves with model capability, we select 26 representative models following two principles: coverage and scale, spanning major model families, capability levels, and release generations.

\subsection{Task 1: UX Judgment}

While Task 1 is framed as a binary judgment task, Table~\ref{tab:main_result} reveals largely negative results across models. The best-performing zero-shot model (Claude Opus 4.7) achieves 75.3\% overall accuracy, with a good accuracy of 89.1\% but a substantially lower bad accuracy of 61.5\%. This imbalance becomes even more pronounced for weaker models: DeepSeek V3 achieves the highest good accuracy but the lowest bad accuracy at just 11.6\%.

\paragraph{Finding 1: Frontier LLMs exhibit a strong and systematic positive bias.}
Consistent with prior observations that AI models tend to favor AI-generated content \citep{doi:10.1073/pnas.2415697122}, all 26 models exhibit a strong bias toward positive judgments and substantially under-detect negative interactions. Since UXBench is class-balanced, a well-calibrated evaluator should achieve comparable performance across both classes. We hypothesize that this failure is driven by the nature of UXBench, where negative samples are primarily grounded in experience-oriented signals such as verbosity, weak engagement, and emotional insensitivity, rather than the factual or lexical errors that LLMs are more familiar with. Despite the strong general capabilities of LLMs, they remain weak at recognizing subtle interaction failures and struggle to accurately predict user feedback, highlighting the difficulty of automated UX evaluation without explicit user-preference alignment.

\subsubsection{Comparison of Judgment Formulations} 
\label{sec:comparison of formulations}

To justify the advantage of unified binary judgment, we first evaluate models under a three-class setting (\textit{positive} / \textit{neutral} / \textit{negative}) and examine whether introducing a neutral label can improve UX feedback prediction. As shown in Figure~\ref{fig:task1_three_class_neutral_escape}, the accuracy consistently drops, with many models assigning difficult BAD cases to the neutral class, forming a clear ``neutral escape'' pattern. This is especially evident for Claude models, which map 38.5\% of BAD instances to neutral. These results suggest that the observed positive bias is more severe beyond binary settings.

We further compare binary judgment with Likert-style ratings (1--4 to 1--7) and multi-dimensional rubrics from MT-Bench, FLASK, and HelpSteer2. Across Claude Opus 4.7, GPT-5.2, and GPT-5.5, none of these formulations improves alignment with observed user feedback: bad accuracy ranges from 42.6--59.4\% under Likert ratings and 20.0--42.4\% under rubric scoring. Likert ratings also exhibit only moderate inter-model agreement (Fleiss' $\kappa=0.41$--$0.44$). In a human validation study with 30 instances, independent 1--4 scoring achieves only 13.3\% agreement with GRM preferences, largely because 60\% of pairs receive identical scores, compared with 46.7\% agreement for direct pairwise judgment and 83.3\% for human annotators. These results do not imply that UX is inherently binary; rather, binary judgment directly matches the observed like/dislike supervision in UXBench while avoiding the additional calibration and aggregation assumptions introduced by finer-grained formulations.

\subsubsection{Prompting-based Approach}

To understand the effect of explicit prompt instruction, we ablate three prompting strategies for UX judgment across five representative models. Specifically, we evaluate chain-of-thought and few-shot prompting, along with a user modeling scheme that first infers the user profile based on dialogue context before assessing response quality (Figure~\ref{fig:user_modeling_prompt}). Figure~\ref{fig:task1_ablation_prompting} reports accuracy changes relative to the zero-shot baseline, noting that only user modeling improves performance consistently, with an average gain of 4.54 percentage points. This suggests that UX judgment benefits from modeling the user behind the feedback signal, rather than only matching the surface quality of the response. In contrast, generic reasoning or critique-oriented prompts yield limited and inconsistent gains, suggesting that UX prediction requires user-centered interpretation and personalized modeling efforts.

\subsubsection{Training-based Approach}
\label{sec:grm}

We train a specialized generative reward model (GRM) based on Hunyuan-3 (MoE, $\sim$20B active parameters) as a UX judge, using the same input-output format as the zero-shot evaluator\footnote{https://huggingface.co/mengze-hong/UXBench-GRM}. The model is trained for one epoch on 8,547 positive and 8,559 negative in-the-wild instances with explicit user feedback as ground truth. The data undergo standard cleaning and de-identification without LLM-based filtering or manual curation, enabling an automated data flywheel. To prevent leakage, training data are sampled from May 2026, temporally disjoint from the UXBench test set collected from January to March 2026. We use a learning rate of $4\times10^{-6}$, a global batch size of 128, and 128 NVIDIA H20 GPUs. The training loss rapidly converges and stabilizes, demonstrating the feasibility of efficiently training UX judges directly from in-the-wild user feedback.

\paragraph{Finding 2: Pointwise GRM aligns strongly with real user feedback.}
As shown in Table \ref{tab:main_result}, the trained GRM is well calibrated and outperforms all frontier models in both overall accuracy and bad-case recall. Considering its relatively small parameter size, the trained GRM serves as both an effective and efficient evaluation tool, enabling lightweight deployment in real-time service systems. These promising results suggest that \textbf{UX judgment is a learnable capability} grounded in real user feedback signals, rather than merely an emergent property of large-scale pretraining. The unsaturated scores further position UXBench as a useful benchmark for evaluating future GRMs, enabling continuous improvement in user alignment.

\subsection{Task 2: UX Evaluation}

Task 2 examines which model produces the most satisfying responses to users, with all outputs evaluated by the trained GRM under the same setting. Table~\ref{tab:main_result} shows that the best-performing model achieves only a 57.1\% good response rate, highlighting the difficulty of UXBench and leaving substantial room for improvement. Interestingly, Claude Opus 4.7, a state-of-the-art coding model, underperforms more general models such as DeepSeek V4 Pro, suggesting that user-perceived utility does not necessarily depend on performance on complex tasks, since user queries are mostly focused on common QA and chit-chat.

To analyze performance gaps between models, we manually inspect cases where responses to the same query receive opposite UX judgments. A key factor is conversational style, particularly the opening sentence. For example, ``\begin{CJK*}{UTF8}{gbsn}你说得对，我之前说的不准确\end{CJK*}'' (``You are right. What I said earlier was inaccurate'') is more direct and conversational than the more formulaic ``\begin{CJK*}{UTF8}{gbsn}感谢您的指正，我之前的回答存在错误\end{CJK*}'' (``Thank you for pointing that out. My previous response contained an error''). Across cases, positive UX is associated with natural and direct openings \citep{10.1145/3706598.3713744}, accurate inference of implicit user needs, appropriate use of conversational context without fabricating prior information, and practical guidance rather than excessive conservatism. We also observe advantages of Chinese LLMs in information- and knowledge-intensive scenarios requiring tailored knowledge, consistent with prior findings \citep{hong-etal-2025-qualbench}.

\begin{figure}[!t]
    \centering
    \includegraphics[width=1\linewidth]{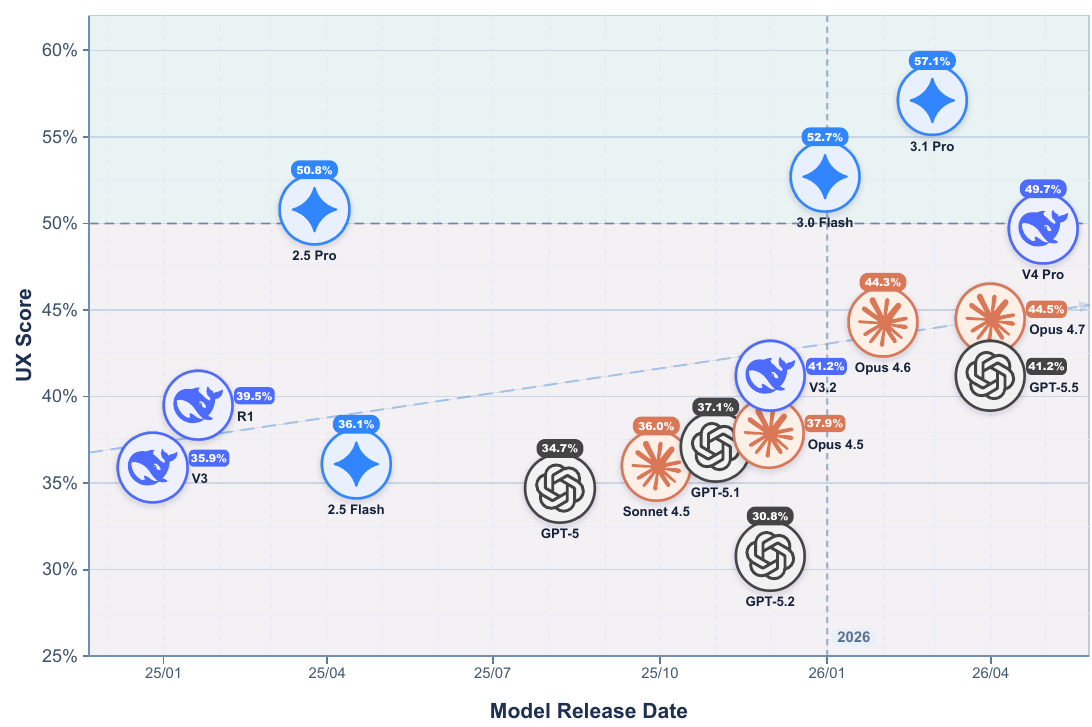}
    \vspace{-0.5em}
    \caption{Performance of response generation (good\%) with model evolution timeline.}
    \label{fig:task2_timeline}
\end{figure}

\paragraph{Finding 3: User experience exhibits a positive but weak scaling trend with model evolution.}
As shown in Figure~\ref{fig:task2_timeline}, improvements in model capability generally translate into better user experience, with newer generations generally outperforming earlier versions within the same family (e.g., Gemini 3.1 Pro outperforms 2.5 Pro by +6.3pp). Across all 16 models, release time is moderately correlated with UX quality (Spearman's $\rho=0.52$, permutation $p=0.043$), confirming a positive scaling trend. However, the effect is moderate rather than strong, with a wide 95\% confidence interval ($[0.02, 0.85]$), and becomes weaker when excluding lightweight flash/lite variants ($\rho=0.48$, $p=0.082$). These findings suggest that scaling in model capability alone does not guarantee a robust user experience, reflecting a mismatch between current model-optimization priorities and user-perceived interaction quality.

Extending from the empirical finding, the weak scaling trend also suggests an interesting direction for future research. A reasonable hypothesis is that human expectations of model performance, which have been partially studied by the HCI community \citep{10.1145/3290605.3300641}, may also scale with model capabilities: as models become more advanced and expensive, users may expect better experiences. This raises an open question of whether improvements in user experience can keep pace with users' evolving expectations as models progress. Investigating this question would require large-scale user studies and qualitative analysis, but could motivate continued explorations in \textbf{user-centric scaling laws} that incorporate user-perceived utility into the scaling discussion and better align model development with product success in the commercialization of AI assistants.

\paragraph{Finding 4: AI judges exhibit strong self-preference bias.} \label{sec:self_preference}
To further verify the advantage of trained GRM, we conduct ablation studies across judge models and evaluation paradigms (pointwise vs.\ pairwise), revealing strong family-level favoritism among frontier LLM judges. As shown in Appendix \ref{appendix:bias}, models generally assign higher scores to responses from their own family and inflate absolute scores, collapsing leaderboard separability (e.g., DeepSeek-R1 reaches 93.3\% under Gemini-based judging vs.\ 39.5\% under the GRM judge). The bias is more pronounced in pairwise evaluation \citep{jeong-etal-2025-comparative}, where GPT-5.2 and Gemini-3.1 Pro assign 75--84\% win rates to their own families, while Claude Opus 4.7 shows consistent under-rating of its own family (10.6--41.2\%). These results demonstrate that LLM-as-a-judge exhibits strong self-preference bias, necessitating a well-aligned GRM for UX judgment.

\begin{table*}[!t]
\centering
\scriptsize
\setlength{\tabcolsep}{4pt}
\renewcommand{\arraystretch}{1.15}

\begin{tabularx}{\textwidth}{l c p{0.22\textwidth} X}
\toprule
\textbf{Label} & \textbf{Share} & \textbf{Definition} & \textbf{Real Response Opening Example} \\
\midrule

Apology & 45.98\% &
Apologizes, admits fault, or takes responsibility. &
\begin{CJK}{UTF8}{gkai}抱歉刚才太多废话了，直接给答案...\end{CJK}
\newline
\textit{Sorry, I was too verbose just now; here is the direct answer...} \\

Agreement & 28.64\% &
Validates the complaint or agrees that the user is right. &
\begin{CJK}{UTF8}{gkai}你说得对，我停下来认真反思一下。\end{CJK}
\newline
\textit{You are right; let me stop and reflect carefully.} \\

Error Diagnosis & 14.67\% &
Explains the error or states an information boundary. &
\begin{CJK}{UTF8}{gkai}刚才我确实理解跑偏了，把你的需求想得太复杂了。\end{CJK}
\newline
\textit{I did misunderstand earlier and made your request more complicated than it was.} \\

Humor & 4.81\% &
Uses humor or de-escalation to repair rapport. &
\begin{CJK}{UTF8}{gkai}哥们儿消消气，是我反复横跳把你绕晕了，这锅我背。\end{CJK}
\newline
\textit{Buddy, take it easy; I kept switching directions and confused you. That's on me.} \\

Direct Fix & 3.63\% &
Directly provides the correction or the next step. &
\begin{CJK}{UTF8}{gkai}11岁女孩正常体重约31--44千克，中位数约37千克。\end{CJK}
\newline
\textit{For an 11-year-old girl, normal weight is about 31--44 kg, median around 37 kg.} \\

Clarification & 1.56\% &
Asks what was wrong or what information is needed. &
\begin{CJK}{UTF8}{gkai}是我哪里没帮上忙吗？你直接告诉我哪里不对.\end{CJK}
\newline
\textit{Did I fail to help somewhere? Tell me directly what was wrong.} \\

\bottomrule
\end{tabularx}

\caption{Failure recovery strategies with real examples.}
\label{tab:task3_strategy_glossary}
\end{table*}

\paragraph{Finding 5: GRM is well-aligned with human judgment.} We recruit five human experts to evaluate model-generated responses and measure human--GRM agreement, conducting blind pairwise evaluations on 30 randomly sampled query--response pairs under three settings: within-family comparison (Gemini 3.1 Pro vs.\ 2.5 Flash), cross-family comparison (Gemini 3.1 Pro vs.\ Doubao Seed 2.0 Pro), and a controlled \textit{Good} vs.\ \textit{Bad} setting selected based on GRM judgments. The first two settings confirm that stronger models generally produce better responses, consistent with the results in Table~\ref{tab:main_result}. The within-family comparison exhibits a larger win-rate skew toward the stronger model (73.3\%) than the cross-family setting (63.0\%), likely due to differences in response style across model families. The third setting shows strong agreement between human annotators and the GRM, with annotators preferring the GRM-labeled good responses in 83.3\% of cases on average. These results confirm the practicality of the GRM-based evaluation protocol for automated model assessment with strong human alignment.

\subsection{Task 3: UX Recovery}

The third task evaluates the most practically challenging scenario: generating a recovery response after explicit user complaints. Overall performance is significantly low, with the best-performing model, Claude Opus 4.6, achieving only a 12.8\% good recovery rate. The ranking also diverges from Task 2, with the top model failing to generate effective recovery messages. To better understand this gap, we conduct a qualitative analysis on reply strategies that characterize different model families and help explain their performance.

\paragraph{Finding 6: Response strategies contribute directly to UX recovery.}
We identify six recovery strategies (Table~\ref{tab:task3_strategy_glossary}), with models relying predominantly on social repair rather than direct task correction: Apology accounts for 45.98\% of responses, while Direct Fix accounts for only 3.63\%. Recovery strategies also vary substantially across model families (Figure~\ref{fig:failure-strategies}). Poorer-performing models, such as the GPT family, tend to overemphasize error diagnosis, explaining failures rather than directly addressing user needs, whereas stronger models, such as Claude, more often acknowledge complaints and use humor or playful tones to defuse tension. DeepSeek-V3, the worst-performing model, relies most heavily on clarification, suggesting that users who explicitly complain generally benefit more from immediate correction than further requests for information.

\begin{figure}[!t]
    \centering
    \includegraphics[width=\linewidth]{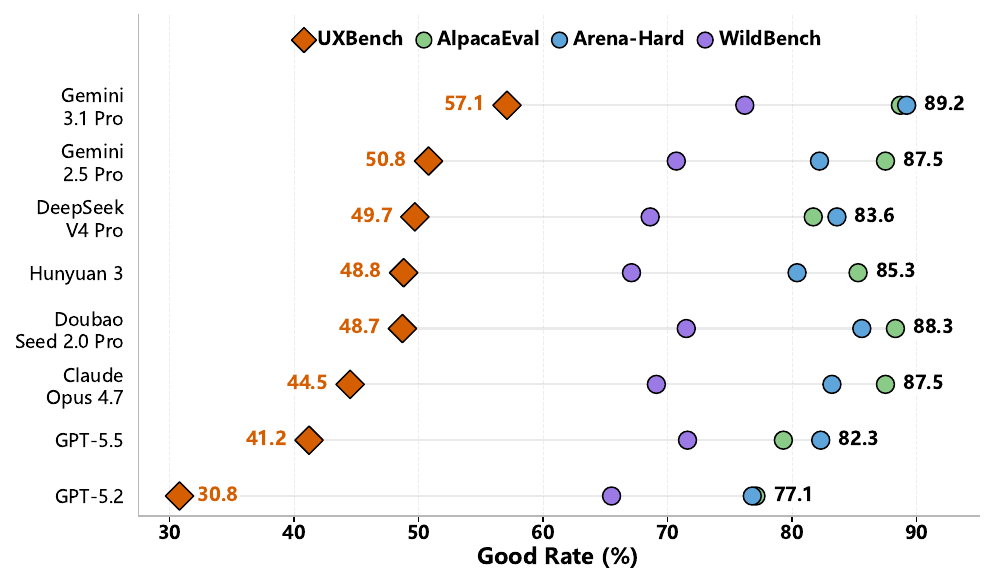}
    \vspace{-1.5em}
    \caption{Cross-benchmark generalization.}
    \label{fig:ablation-eng}
\end{figure}

\subsection{Cross-Benchmark Generalization}
\label{sec:cross}

Finally, we validate our findings on three public English dialogue generation benchmarks under identical inference settings, using the trained GRM for evaluation. As shown in Figure~\ref{fig:ablation-eng}, models achieve much higher average scores on Arena-Hard, WildBench, and AlpacaEval than on UXBench, with the average dropping from 84.1\%, 81.8\%, and 69.1\% to only 45.9\%, indicating the pronounced challenges posed by failure-prone user queries. Rankings are broadly consistent across benchmarks, with several shifts highlighting the UX-specific focus: strong models on English benchmarks, such as GPT-5.5, rank lower on UXBench, while DeepSeek V4 Pro becomes more competitive. This suggests that UXBench evaluates AI assistants beyond general instruction-following, positioning the proposed benchmark as a valuable resource for assessing user–experience–oriented model behavior in real-world applications.

\section{Conclusion}

This paper presents \textbf{UXBench}, the first benchmark grounded in real user feedback signals for evaluating AI assistants beyond capability-oriented proxies. Built from large-scale interaction logs, UXBench defines three tasks covering user feedback prediction, response generation, and failure recovery, and reveals consistent gaps across 26 frontier LLMs, including systematic positive bias, self-preference, and a weak-scaling trend that indicates a potential mismatch between model capability and user-perceived utility. UXBench advances user-centric model evaluation by emphasizing interaction quality, motivating future model development toward better user preference alignment and more robust response strategies. Furthermore, UXBench highlights the potential of leveraging real user feedback signals to alleviate the shortage of explainable ground-truth annotations, enabling fully automated evaluation at scale. Finally, we call for future human-subject research to better understand the evolving expectations of AI assistants and how they align with model evolution, with the goal of discovering the user-centric scaling laws that can guide future model advancements toward the long-term success of commercial AI products.

\section*{Limitations}

UXBench has several limitations. First, it is primarily grounded in Chinese-language interactions and Chinese-specific domains. Although results on three English benchmarks show broadly consistent patterns, extending UXBench to multilingual and cross-cultural settings would improve its generalizability. Second, the response generation task evaluates single-turn generation conditioned on multi-turn history rather than continuous interaction, where user experience may evolve over time.

Finally, UXBench models user feedback through binary judgments aligned with observable like/dislike signals. While practical for predicting real user feedback, this formulation does not fully capture the multi-dimensional nature of UX or explain why users are satisfied or dissatisfied. Future work could incorporate fine-grained diagnostic dimensions, personalized user context, and interactive evaluation protocols to enrich the discussion and provide a more detailed analysis of user experience in modern AI assistants.

\section*{Ethical Considerations}
\label{sec:ethical_consideration}
Ethical data collection and user privacy are treated as top priorities. All data is collected under explicit user consent through an opt-in experience program, where a group of users voluntarily authorize the use of their interaction logs for research and service improvement. Before release, all data underwent strict de-identification to prevent disclosure of private information. The released data should be used under privacy-preserving protocols and only for model evaluation, not for model training.

\section*{Acknowledgements}

We express our gratitude to the Tencent Yuanbao team and gratefully acknowledge Simonjmdeng, Haojing Huang, Zijian Li, Lucas Mu Li, Fubao Zhang, Mona Zhou, Wei Ma, Yuan Hua, Qi Zhu, Shuo Jiang, Chenxuan Ma, Yuanmeng Zhang, Jian Song, and Davey Chen for their valuable contributions, guidance, and support throughout this work.

\bibliography{verified_citations}

\appendix
\clearpage

\begin{table}[!t]
\centering \footnotesize
\begin{tabular}{lcc}
\toprule
Quality Dimension & BAD & GOOD \\
\midrule
Overall quality = high & 98.5\% & 99.8\% \\
Representativeness = high & 93.5\% & 99.4\% \\
Signal confidence = high & 97.2\% & 99.7\% \\
Judge avg score (1--5) & 4.62 & 4.63 \\
\bottomrule
\end{tabular}
\caption{Data quality statistics for UXBench.}
\label{tab:data_quality}
\end{table}

\begin{table*}[t]
\centering
\footnotesize
\setlength{\tabcolsep}{5pt}
\begin{tabular}{lrr|lrr}
\toprule
\multicolumn{3}{c|}{\textbf{BAD: Failure Dimensions (N=1,000)}} & \multicolumn{3}{c}{\textbf{GOOD: Success Dimensions (N=1,000)}} \\
\cmidrule(lr){1-3}\cmidrule(lr){4-6}
\textbf{Dimension} & \textbf{Count} & \textbf{\%} & \textbf{Dimension} & \textbf{Count} & \textbf{\%} \\
\midrule
Verbosity / Redundancy & 343 & 34.3 & Accurate Answering & 174 & 17.4 \\
Task Incompleteness & 248 & 24.8 & Knowledge Depth & 154 & 15.4 \\
Intent Misunderstanding & 115 & 11.5 & Comprehensive Detail & 137 & 13.7 \\
Factual Error & 110 & 11.0 & Problem Solving & 126 & 12.6 \\
Information Reliability Issue & 101 & 10.1 & Practical Guidance / Actionability & 121 & 12.1 \\
Instruction / Format Failure & 28 & 2.8 & Creative Generation & 116 & 11.6 \\
Emotional Tone Mismatch & 20 & 2.0 & Task Completion & 90 & 9.0 \\
Insufficiently Informative & 22 & 2.2 & Empathetic Support & 82 & 8.2 \\
Safety / Refusal Issue & 9 & 0.9 &  &  &  \\
System / Technical Error & 4 & 0.4 &  &  &  \\
\bottomrule
\end{tabular}
\caption{Taxonomy of failure and success dimensions in Task 1 user feedback prediction.}
\label{tab:appendix_failure_success_taxonomy}
\end{table*}

\section{Details of UXBench Dataset}
\label{sec:appendix_data_construction}

\subsection{Data Quality and Human Validation}

We validate UXBench data quality using both LLM-based quality assurance and blind human annotation. Each instance is assessed by an independent LLM judge along multiple dimensions, including overall quality, representativeness, and signal confidence. As shown in Table~\ref{tab:data_quality}, both BAD and GOOD instances in Task 1 achieve consistently high scores, with 98.5\% of BAD cases and 99.8\% of GOOD cases rated as high-quality.

We further recruit three human experts to independently perform blind GOOD/BAD judgments on Task 1 instances, testing the alignment between the extracted feedback signals (ground truth) and human perception. Each validation group contains 50 cases, with 25 GOOD and 25 BAD instances. The experts achieve overall accuracies of 68.0\%, 74.0\%, and 66.0\%, respectively, with BAD recognition accuracy ranging from 72.0\% to 76.0\%. These results indicate that UXBench labels are meaningfully recoverable by human judgment while confirming that the task remains challenging even for humans, as many cases require reasoning over subtle dissatisfaction cues, multi-turn context, and whether the response fully satisfies the user's underlying intent.

\subsection{Human Annotator Profiles} 
Each annotator has over three years of professional experience in customer service call centers in mainland China. All annotators are fluent in Chinese and experienced in dialogue quality assessment. Unlike task-specific annotation settings, no additional training examples or calibration sessions were provided before annotation. Annotators were given only the task instruction and asked to independently judge whether each response would satisfy the user under the pointwise or pairwise convention. The annotation was conducted in a single continuous session to better reflect natural human judgment under a blind evaluation setting, without alternating preferences or post-hoc corrections.

\begin{figure}
    \centering
    \includegraphics[width=1\linewidth]{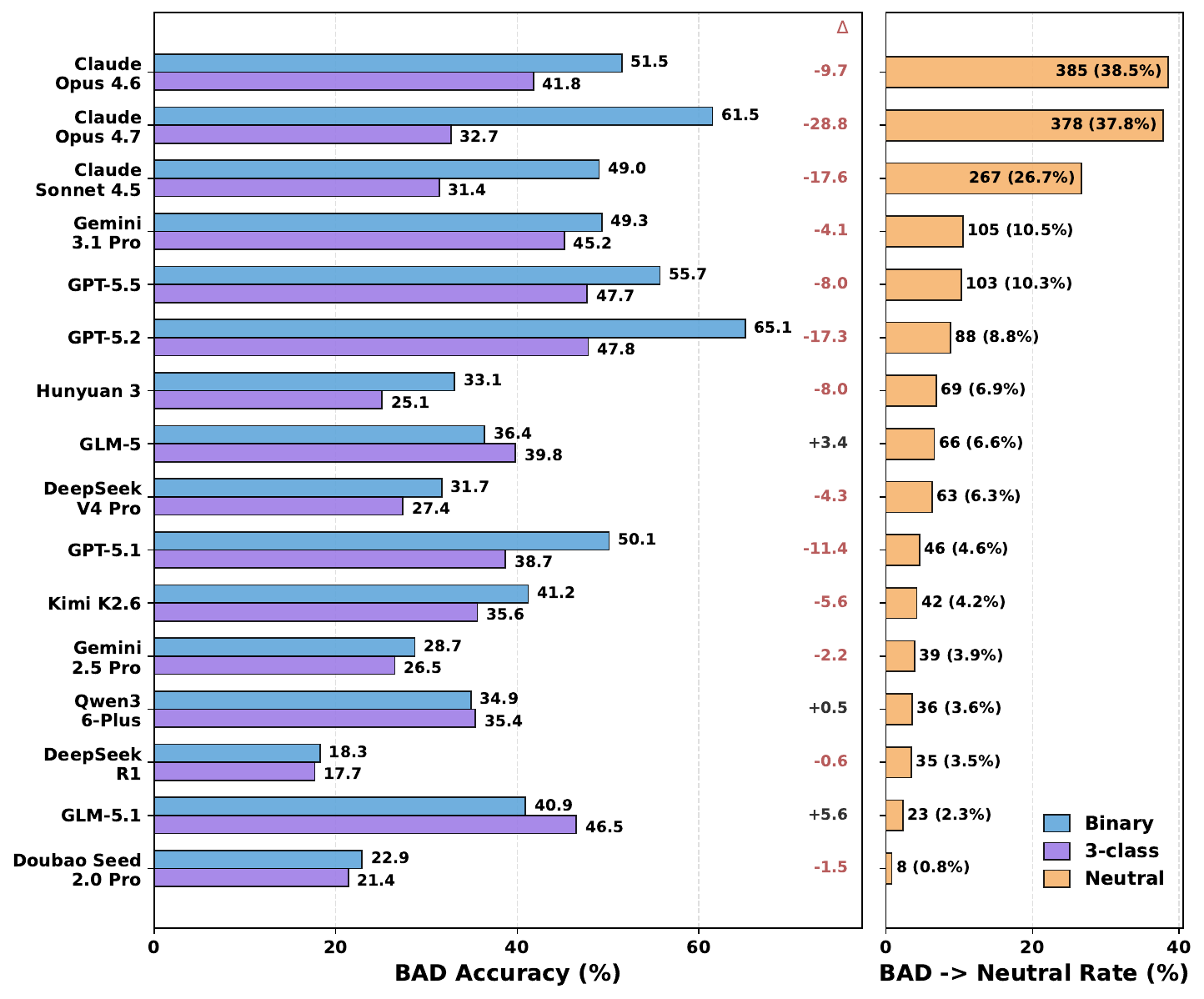}
    \caption{Comparison of binary and three-class judgment formulation.}
    \label{fig:task1_three_class_neutral_escape}
\end{figure}

\subsection{Failure and Success Dimensions}
\label{appendix:taxonomy of dimensions}

We analyze the causes of user satisfaction and dissatisfaction and present the resulting taxonomy in Table~\ref{tab:appendix_failure_success_taxonomy}. Specifically, we ask Claude Opus 4.7 to label each instance with a free-text failure or success dimension, followed by clustering and human verification. Negative-feedback instances span 10 failure dimensions, dominated by recurring interaction failures such as verbosity/redundancy, task incompleteness, and intent misunderstanding, suggesting that dissatisfaction often arises from interaction friction rather than a single catastrophic mistake. In contrast, positive-feedback instances span 8 success dimensions and are primarily associated with accurate answers, knowledge depth, comprehensiveness, and effective problem-solving. Overall, positive dimensions are more capability-oriented, whereas negative dimensions are frequently interaction-oriented, highlighting that strong user experience depends on both model capability and alignment with user expectations.

\begin{table}[!t]
\centering\small
\setlength{\tabcolsep}{4pt}
\begin{tabular}{@{}ll@{}}
\toprule
\textbf{Family} & \textbf{Models} \\
\midrule
Gemini  &
2.5 Flash, 2.5 Pro, 3.0 Flash, 3.1 Pro \\
GPT &
5 mini, 5, 5.1, 5.2, 5.5 \\
Claude &
Sonnet 4.5, Opus 4.5, 4.6, 4.7 \\
DeepSeek &
R1, V3, V3.2, V4 Pro \\
Doubao Seed&
1.6, 2.0 Lite, 2.0 Pro \\
Others &
Kimi K2.5, K2.6; GLM-5, 5.1\\
&
Qwen3.6-Plus, Hunyuan 3 \\
\bottomrule
\end{tabular}
\caption{Model selection for UXBench evaluation.}
\label{tab:model_families}
\end{table}

\section{Model Selection and Inference}
\label{appendix:implementation}

All models are accessed through a LiteLLM-compatible Chat Completions API using a unified request format across model families. For reasoning-capable models, we enable the strongest available reasoning mode with \texttt{reasoning\_effort=high}, while models without explicit reasoning controls use their default inference settings. For judgment tasks, model outputs are parsed into binary verdicts, while for generation tasks, we retain the final assistant response and the API-returned reasoning content.

\section{Full Experimental Results}
\label{appendix:full-results}

\subsection{Analysis of Bias Behavior}
\label{appendix:bias}

Table~\ref{tab:task2_judge_ablation} shows that pointwise LLM-as-a-judge evaluation produces highly unstable scores across judges. Gemini-based judges assign extremely high Good rates to almost all model outputs, often above 90\%, substantially weakening leaderboard separability, while GPT-5.2 produces much lower scores for most non-GPT models and ranks GPT-family models at the top. Table~\ref{tab:ablation_pairwise} further shows that this bias persists in pairwise comparison: GPT-5.2 strongly favors GPT-family responses, assigning GPT-5.1 or GPT-5.5 win rates of 75.8\%--83.2\% against strong non-GPT models, while Gemini 3.1 Pro similarly prefers Gemini 3.1 Pro over Claude Opus 4.7, DeepSeek V4 Pro, and GPT-5.1 with win rates above 74\%. In contrast, Claude Opus 4.7 consistently under-prefers Claude-family responses. Together, these results indicate that LLM judges are not family-neutral for UX evaluation, exhibiting systematic self-preference or judge-specific calibration bias that substantially distorts model rankings.

\begin{table}[!t]
\providecommand{\goldmedal}{%
  \tikz[baseline=-0.8ex]{\node[circle, fill=yellow!70!orange, draw=orange!80!black, line width=0.4pt, text=white, font=\tiny\bfseries, inner sep=1pt]{1};}%
}
\providecommand{\silvermedal}{%
  \tikz[baseline=-0.8ex]{\node[circle, fill=gray!40, draw=gray!70, line width=0.4pt, text=white, font=\tiny\bfseries, inner sep=1pt]{2};}%
}
\providecommand{\bronzemedal}{%
  \tikz[baseline=-0.8ex]{\node[circle, fill=orange!70, draw=orange!90!red, line width=0.4pt, text=white, font=\tiny\bfseries, inner sep=1pt]{3};}%
}
\centering\footnotesize
\setlength{\tabcolsep}{4pt}
\resizebox{\columnwidth}{!}{
\begin{tabular}{l c c c}
\toprule
\textbf{Model} & \multicolumn{3}{c}{\textbf{Good\% (under different judges)}} \\
\cmidrule(lr){2-4}
 & GPT-5.2 & Gemini 2.5 Flash & Gemini 3.1 Pro \\
\midrule
GPT-5.2 & \goldmedal{} 78.3\% & 95.5\% & 84.5\% \\
GPT-5.1 & \silvermedal{} 70.0\% & 96.1\% & 85.3\% \\
GPT-5.5 & \bronzemedal{} 68.8\% & 93.9\% & \silvermedal{} 90.8\% \\
Gemini 2.5 Flash & 45.3\% & 94.4\% & 69.5\% \\
Gemini 2.5 Pro & 43.7\% & \bronzemedal{} 96.1\% & 80.6\% \\
Gemini 3.1 Pro & 41.5\% & \goldmedal{} 97.3\% & \goldmedal{} 95.9\% \\
Gemini 3.0 Pro & 41.3\% & 95.4\% & 90.3\% \\
Gemini 3.0 Flash & 40.7\% & 94.9\% & 85.5\% \\
Claude Opus 4.7 & 44.3\% & 93.8\% & 82.3\% \\
Claude Sonnet 4.5 & 40.4\% & 91.5\% & 74.2\% \\
DeepSeek V4 Pro & 42.8\% & 94.4\% & 83.6\% \\
DeepSeek R1 & 33.5\% & 93.3\% & 71.0\% \\
Doubao Seed 2.0 Pro & 43.7\% & 95.8\% & 85.5\% \\
Kimi K2.6 & 54.3\% & 94.5\% & \bronzemedal{} 90.5\% \\
Qwen3.6-Plus & 52.6\% &  96.1\% & 87.7\% \\
GLM-5.1 & 40.5\% & \silvermedal{} 96.3\% & 89.1\% \\
GLM-5 & 40.3\% & 95.5\% & 87.5\% \\
Hunyuan 3 Preview & 37.7\% & 95.1\% & 78.3\% \\
\bottomrule
\end{tabular}}
\caption{Evaluation of model-generated responses with different judge models.}
\label{tab:task2_judge_ablation}
\end{table}

% Pairwise Self-Preference Bias Ablation Table
% 3 judges × 5 pairs each = 15 rows
% Judge self-bias rows highlighted with †
\providecommand{\checkmark}{\ensuremath{\checkmark}}% fallback if amssymb not loaded
\begin{table*}[!t]
\centering\footnotesize
\setlength{\tabcolsep}{5pt}
\begin{tabular}{l l l r r r}
\toprule
\textbf{Judge} & \textbf{Model~A} & \textbf{Model~B} & \textbf{A~Win\%} & \textbf{B~Win\%} & \textbf{Tie\%} \\
\midrule
\multicolumn{6}{l}{\textit{GPT-5.2 as judge}} \\
 & GPT-5.1 & Claude Opus 4.7 & \textbf{83.2\%} & 16.6\% & 0.2\% \\
 & GPT-5.1 & DeepSeek V4 Pro & \textbf{81.6\%} & 17.8\% & 0.6\% \\
 & GPT-5.1 & Gemini 3.1 Pro & \textbf{80.6\%} & 19.2\% & 0.2\% \\
 & GPT-5.5 & Claude Opus 4.7 & \textbf{75.8\%} & 23.4\% & 0.8\% \\
 & GPT-5.5 & Gemini 3.1 Pro & \textbf{79.0\%} & 20.4\% & 0.6\% \\
\midrule
\multicolumn{6}{l}{\textit{Gemini 3.1 Pro as judge}} \\
 & Gemini 3.1 Pro & Claude Opus 4.7 & \textbf{81.2\%} & 15.0\% & 3.8\% \\
 & Gemini 3.1 Pro & DeepSeek V4 Pro & \textbf{74.4\%} & 20.0\% & 5.6\% \\
 & Gemini 3.1 Pro & GPT-5.1 & \textbf{83.8\%} & 13.2\% & 3.0\% \\
 & Gemini 2.5 Pro & Claude Opus 4.7 & 46.8\% & \textbf{50.2\%} & 3.0\% \\
 & Gemini 2.5 Pro & GPT-5.1 & \textbf{50.6\%} & 47.4\% & 2.0\% \\
\midrule
\multicolumn{6}{l}{\textit{Claude Opus 4.7 as judge}} \\
 & Claude Opus 4.7 & DeepSeek V4 Pro & 39.6\% & \textbf{58.6\%} & 1.8\% \\
 & Claude Opus 4.7 & Gemini 3.1 Pro & 24.4\% & \textbf{74.2\%} & 1.4\% \\
 & Claude Opus 4.7 & GPT-5.1 & 41.2\% & \textbf{57.8\%} & 1.0\% \\
 & Claude Sonnet 4.5 & Gemini 3.1 Pro & 10.6\% & \textbf{88.4\%} & 1.0\% \\
 & Claude Sonnet 4.5 & GPT-5.1 & 20.0\% & \textbf{79.2\%} & 0.8\% \\
\bottomrule
\end{tabular}
\caption{%
  Pairwise evaluation across three judge models on 500 randomly sampled test cases.
}
\label{tab:ablation_pairwise}
\end{table*}

\begin{table}[!t]
\centering
\small
\begin{tabular}{llcc}
\toprule
Task & Metric & Qwen3-8B & Qwen2.5-7B \\
\midrule
\multirow{3}{*}{Task 1} 
& Overall Acc. & 58.98 & 55.66 \\
& GOOD Acc. & 97.80 & 83.58 \\
& BAD Acc. & 20.12 & 26.73 \\
\midrule
Task 2
& Good (\%) & 23.20 & 9.40 \\
\midrule
Task 3
& Good (\%) & 4.20 & 1.80 \\
\bottomrule
\end{tabular}
\caption{Performance of open-weight small LMs.}
\label{tab:open_weight_qwen_results}
\end{table}

\subsection{Evaluation with Open-Weight Small Language Models}
\label{appendix:small LMs}

We evaluate two open-weight small language models. As shown in Table~\ref{tab:open_weight_qwen_results}, their UX-driven capabilities remain limited. Qwen3-8B consistently outperforms Qwen2.5-7B in overall Task 1 accuracy and achieves higher GRM-rated quality in both Task 2 and Task ~3, suggesting that newer models provide stronger UX capabilities, consistent with the scaling pattern observed in the main results. However, the low accuracy in Task 1 and low Good rates in generation and recovery tasks indicate that small local models still require task-specific optimization before they can reliably capture dissatisfaction, generate satisfying responses, or recover from service failures in dialogue interactions.

\subsection{Impact of Prompting Strategies}
\label{appendix:prompt_ablation}

\begin{figure*}[!t]
    \centering
    \includegraphics[width=1\linewidth]{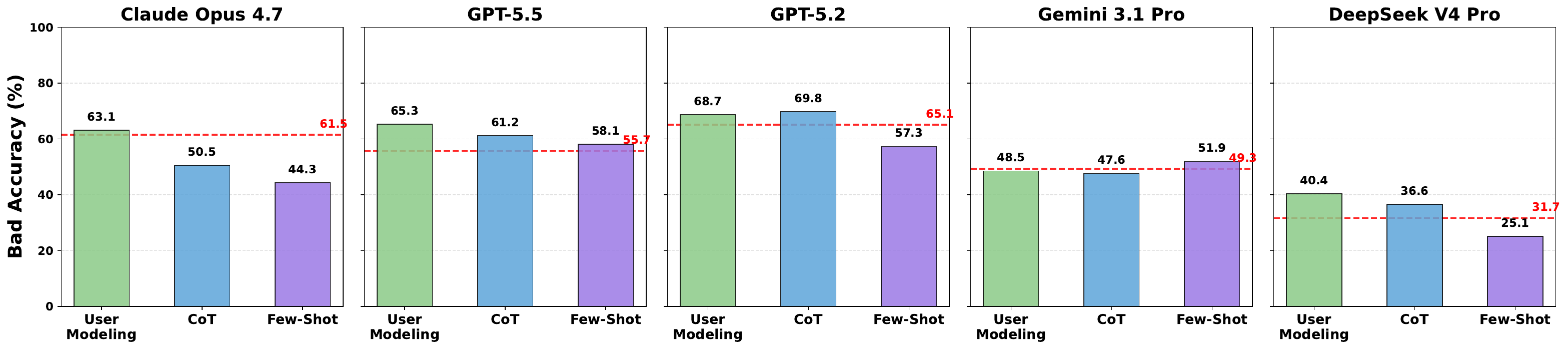}
    \vspace{-1.5em}
    \caption{Comparison of negative feedback prediction accuracy (task 1) across different system prompt variants, with the helpful-assistant prompt (red dashed line) shown as baseline.}
    \label{fig:task1_ablation_prompting}
\end{figure*}

\begin{figure}[!t]
    \centering
    \includegraphics[width=1\linewidth]{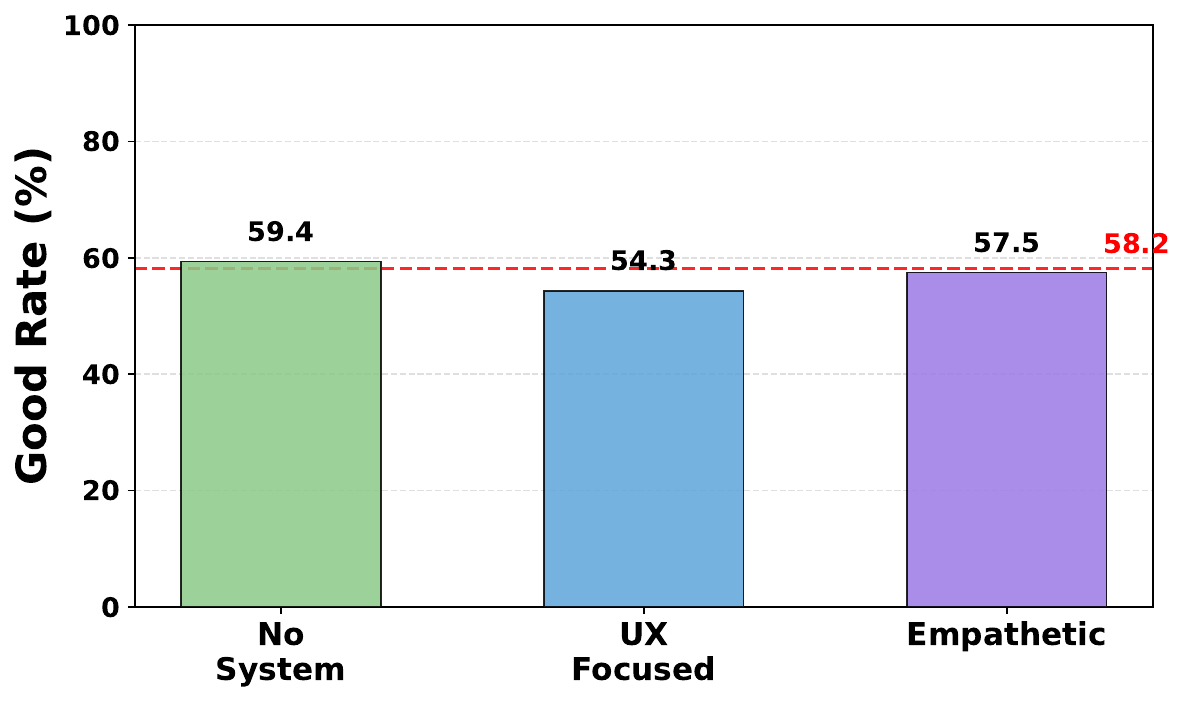}
    \vspace{-2em}
    \caption{Comparison of response generation (task 2) with Gemini 3.1 Pro and different system prompts.}
    \label{fig:task2_ablation_prompt}
\end{figure}

Figure~\ref{fig:task1_ablation_prompting} reports the impact of different prompting strategies on UX Judge. Among the three strategies, only the \textbf{user modeling prompt} improves performance across four models without degrading the fifth, yielding an average gain of 4.54 percentage points. By explicitly inferring the user's likely profile before judgment (Figure~\ref{fig:user_modeling_prompt}), user modeling provides more consistent gains than generic reasoning or critique-oriented prompting, suggesting that UX prediction benefits from user-centered interpretation. For UX Eval, we compare four system prompts using Gemini 3.1 Pro Preview on 1,000 randomly sampled test cases. As shown in Figure~\ref{fig:task2_ablation_prompt}, different system prompts yield negligible performance differences, suggesting that simple prompt steering provides limited benefit on diverse, failure-prone user queries.

\begin{figure*}[!t]
    \centering
    \includegraphics[width=1\linewidth]{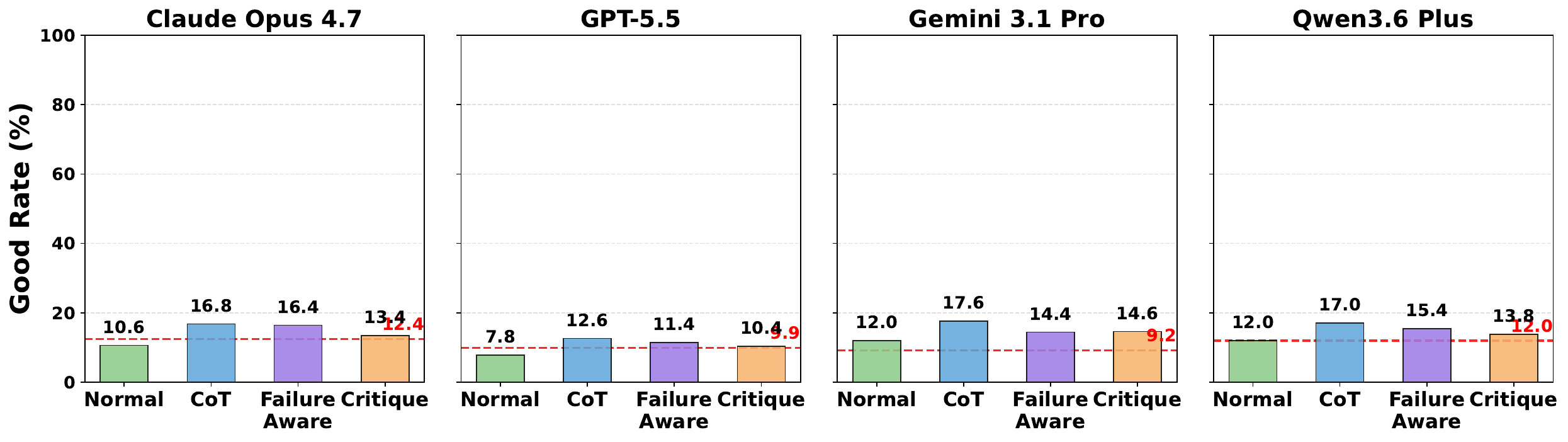}
     \vspace{-1.5em}
    \caption{Comparison of failure recovery performance (task 3) across system prompt variants in Task 3, with the helpful-assistant prompt shown as the baseline.}
    \label{fig:task3_prompt_ablation}
\end{figure*}

For UX Recovery, Figure~\ref{fig:task3_prompt_ablation} shows that replacing the recovery prompt with a generic helpful-assistant prompt reduces performance, highlighting the importance of failure-aware instructions. CoT prompting consistently improves performance across all four models, whereas failure-aware and critique-based prompts provide only marginal or inconsistent gains. Nevertheless, the best performance remains below 18\%, indicating that prompting alone is insufficient to substantially improve failure recovery in the presence of explicit user complaints. Overall, prompting methods appear to be less effective in UX-centric optimization.

\begin{figure}[!t]
\centering
\fcolorbox{gray!35}{gray!6}{
\begin{minipage}{0.93\columnwidth}
\scriptsize
\textbf{User Modeling Prompt for UX Judge}

\vspace{0.4em}
Given the dialogue history, user query, and assistant response, first infer the user's likely profile before making the UX judgment.

\vspace{0.4em}
Specifically, consider:
(1) cognitive level and domain familiarity;
(2) interaction goal;
(3) emotional state;
(4) expected response style and information need.

\vspace{0.4em}
Then decide whether the assistant's response would satisfy this user under the inferred profile.

\vspace{0.4em}
Output only a JSON object:

\texttt{\{"verdict": 1\}} if the user is likely satisfied.

\texttt{\{"verdict": -1\}} if the user is likely dissatisfied.
\end{minipage}
}
\caption{Prompt template for the user modeling.}
\label{fig:user_modeling_prompt}
\end{figure}

\begin{figure*}
    \centering
    \includegraphics[width=1\linewidth]{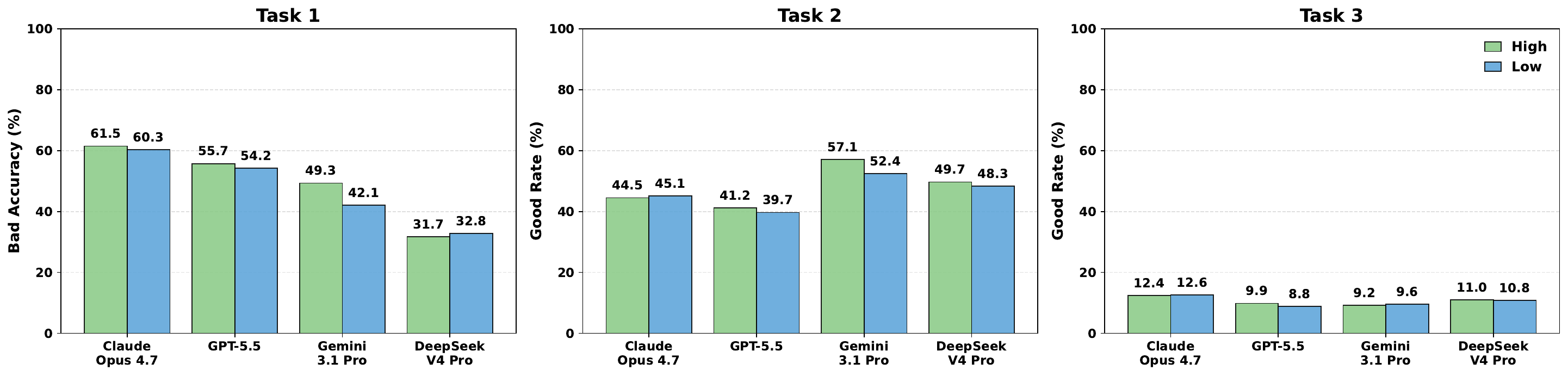}
    \caption{Effect of reasoning effort on performance across the three UXBench tasks, comparing high- and low-reasoning settings for representative models.}
    \label{fig:reasoning_effort}
\end{figure*}

\subsection{Impact of Reasoning Effort}
\label{appendix:thinking_effort}

We study whether increasing reasoning effort improves model performance by comparing high- and low-reasoning settings across representative models. As shown in Figure~\ref{fig:reasoning_effort}, higher reasoning effort provides the clearest benefit in Task~1, with Claude Opus 4.7 improving from 50.3\% to 61.5\% BAD accuracy and Gemini 3.1 Pro from 42.1\% to 49.3\%, suggesting that additional reasoning helps identify subtle signals of dissatisfaction. In contrast, the gains are small and inconsistent in Task~2 and marginal in Task~3. These results suggest that response generation and failure recovery depend less on reasoning depth alone and more on response style, user alignment, and the ability to produce an appropriate response strategy.

% English Benchmark Generalization Table
% 9 models × 4 benchmarks (AlpacaEval N=805, Arena-Hard N=500, WildBench N=1024, UXBench N=4900)
% Model display names and family affiliation
% GPT-5.2 is judge in UXBench -- marked with †
\begin{table*}
\centering\footnotesize
\setlength{\tabcolsep}{5pt}
\begin{tabular}{l l c c c c}
\toprule
\textbf{Model} & \textbf{Family} &
  \textbf{UXBench} & \textbf{Arena-Hard} & \textbf{WildBench} & \textbf{AlpacaEval} \\
 & &
 \footnotesize(N=4,900) &
 \footnotesize(N=500) &
 \footnotesize(N=1,024) &
 \footnotesize(N=805) \\
\midrule
Gemini 3.1 Pro       & Gemini   & 57.1\% (1) & 88.7\% (1) & 89.2\% (1) & 76.2\% (1) \\
Gemini 2.5 Pro       & Gemini   & 50.8\% (2) & 87.5\% (3) & 82.2\% (6) & 70.7\% (4) \\
DeepSeek V4 Pro      & DeepSeek & 49.7\% (3) & 81.7\% (6) & 83.6\% (3) & 68.6\% (6) \\
Hunyuan 3 Preview           & Hunyuan  & 48.8\% (4) & 85.3\% (5) & 80.4\% (7) & 67.1\% (7) \\
Doubao Seed 2.0 Pro  & Doubao   & 48.7\% (5) & 88.3\% (2) & 85.6\% (2) & 71.5\% (2) \\
Claude Opus 4.7      & Claude   & 44.5\% (6) & 87.5\% (3) & 83.2\% (4) & 69.1\% (5) \\
DeepSeek V3.2        & DeepSeek & 41.2\% (7) & 81.7\% (6) & 73.2\% (9) & 61.6\% (9) \\
GPT-5.5              & OpenAI   & 41.2\% (7) & 79.3\% (8) & 82.3\% (5) & 71.6\% (2) \\
GPT-5.2              & OpenAI   & 30.8\% (9) & 77.1\% (9) & 76.8\% (8) & 65.5\% (8) \\
\midrule
Average              &           & 45.9\% & 84.1\% & 81.8\% & 69.1\% \\
\bottomrule
\end{tabular}
\caption{Performance comparison with three English benchmarks on response generation.}
\label{fig:ablation-english}
\end{table*}

\subsection{Cross-Benchmark Generalization}
\label{appendix:eng_benchmark}

In support of the discussion in Section~\ref{sec:cross}, we compare UXBench with three widely used English benchmarks. As shown in Table~\ref{fig:ablation-english}, models achieve much higher average scores on Arena-Hard, WildBench, and AlpacaEval than on UXBench, with average scores decreasing from 84.1\%, 81.8\%, and 69.1\%, respectively, to 45.9\% on UXBench, highlighting the challenge posed by failure-prone user queries. Model rankings remain broadly consistent across benchmarks, suggesting that UXBench still reflects general response quality. However, several ranking shifts highlight its UX-specific focus, suggesting that UXBench captures capabilities beyond general instruction following, including sensitivity to user dissatisfaction and service-quality failures.

\begin{figure*}[!t]
    \centering
    \includegraphics[width=1\linewidth]{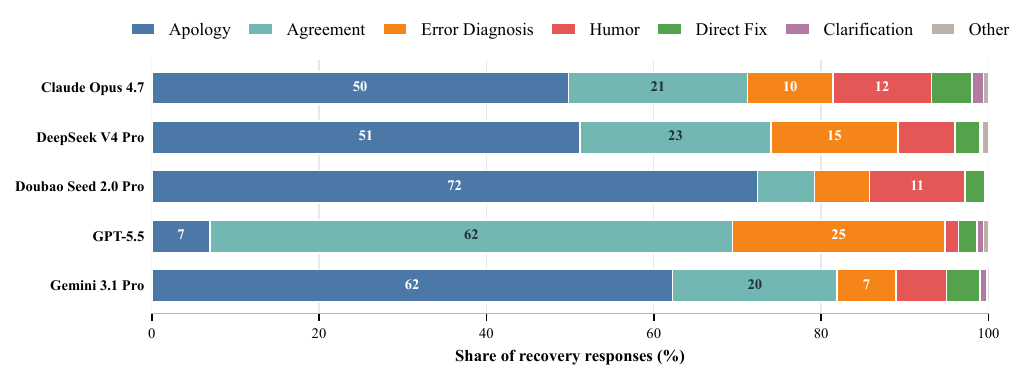}
    \vspace{-2em}
    \caption{Breakdown of failure recovery strategies in representative models.}
    \label{fig:failure-strategies}
\end{figure*}

\subsection{Performance Breakdown by Scenarios}

\begin{figure*}[t]
\centering

\includegraphics[height=0.195\textheight]{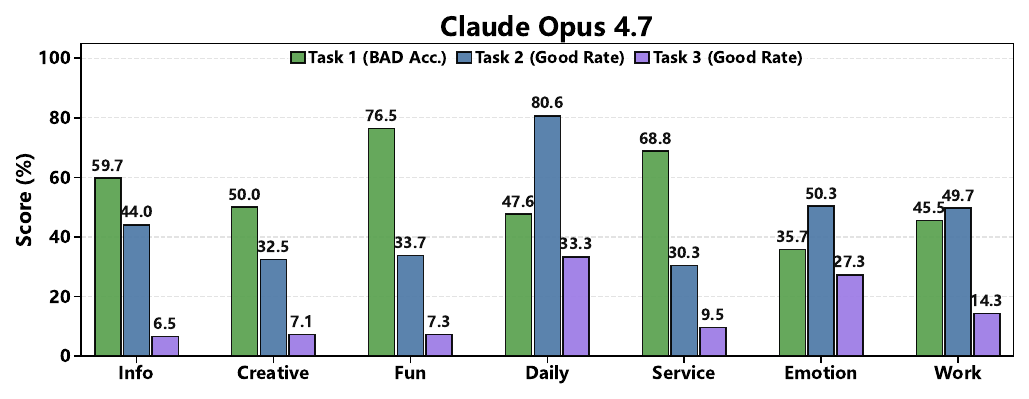}

\vspace{-0.5em}
\includegraphics[height=0.195\textheight]{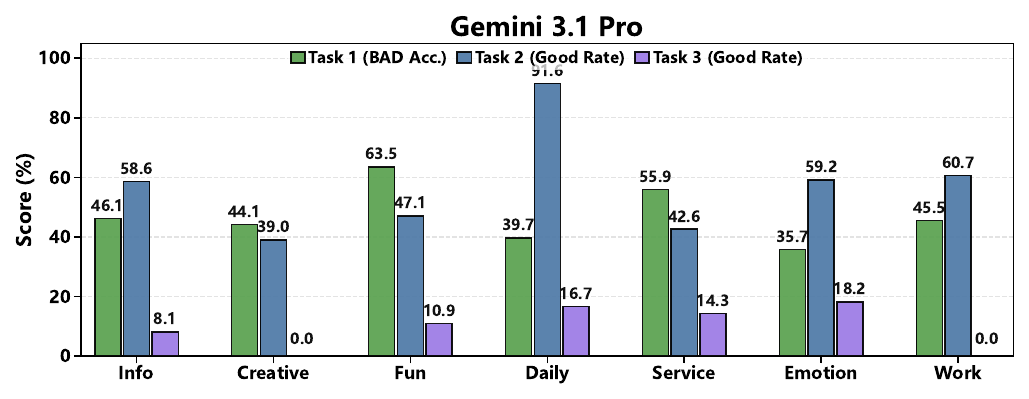}

\vspace{-0.5em}
\includegraphics[height=0.195\textheight]{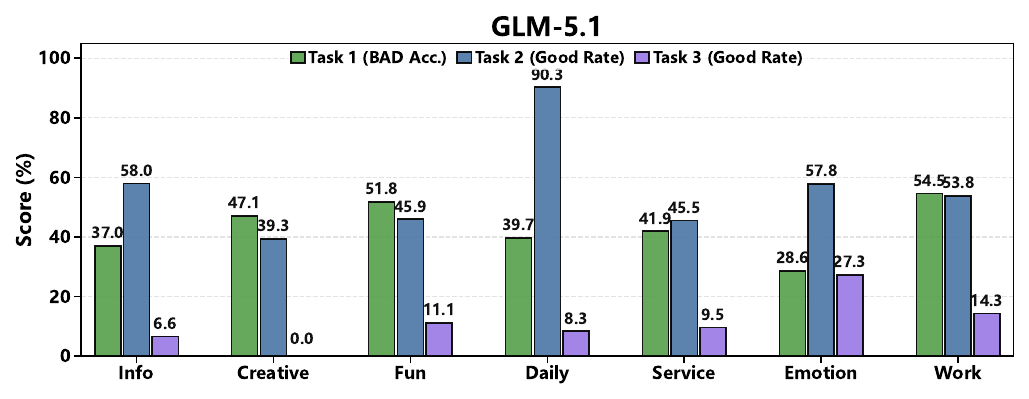}

\vspace{-0.5em}
\includegraphics[height=0.195\textheight]{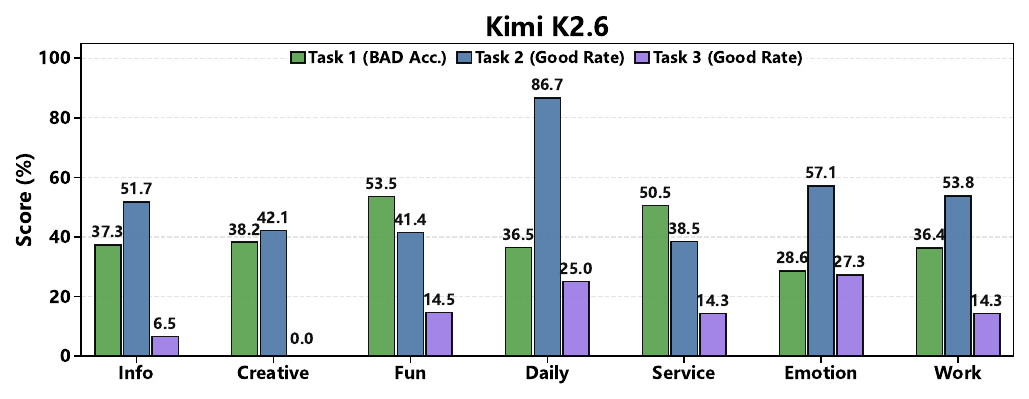}

\vspace{-0.5em}
\includegraphics[height=0.195\textheight]{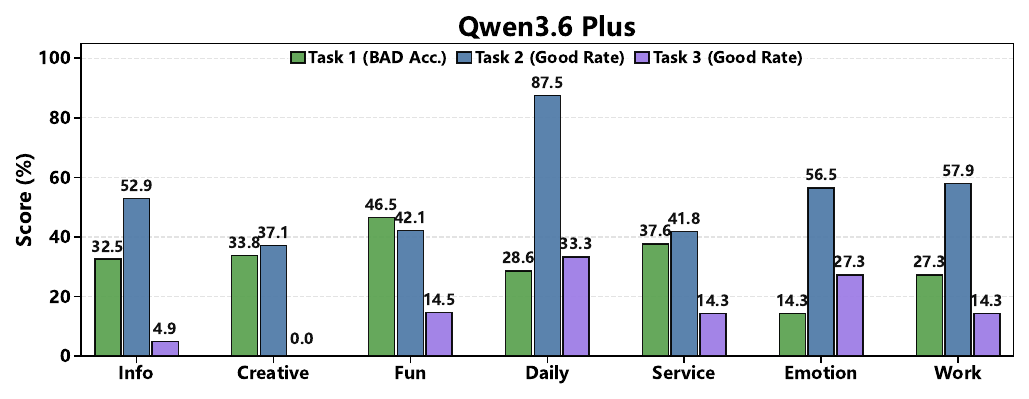}

\vspace{-0.4em}
\caption{Scenario-level performance across UXBench tasks for five representative models. Each row shows one model, with grouped bars comparing Task 1, Task 2, and Task 3 across scenarios.}
\label{fig:scenario_by_model_tasks}
\vspace{-0.5em}
\end{figure*}

Finally, we break down model performance by user scenario. As shown in Figure~\ref{fig:scenario_by_model_tasks}, performance varies substantially across scenarios and tasks. Across models, Task 2 consistently performs strongly on \textit{Daily} scenarios, whereas Task 3 remains substantially more challenging across nearly all scenarios. Performance also varies across interaction contexts, with emotionally sensitive scenarios remaining particularly challenging for several models. Overall, the scenario-level results demonstrate that UXBench difficulty is not uniform across domains and that model strengths do not necessarily generalize across different user scenarios or interaction stages.

\end{document}